\documentclass{article}%
\usepackage[T1]{fontenc}%
\usepackage[utf8]{inputenc}%
\usepackage{lmodern}%
\usepackage{textcomp}%
\usepackage{lastpage}%
\usepackage{array}
\usepackage{geometry}%
\usepackage{multicol}%
\usepackage{graphicx}%
\usepackage{lipsum}%
\usepackage{caption}%
\usepackage{booktabs}%
\usepackage{ragged2e}%
\usepackage{breqn}  % Pour avoir des équations dmath sur 2 lignes.

\title{\textit{ck-means}, a novel unsupervised learning method that combines fuzzy and crispy clustering methods to extract intersecting data}%
\author{Jean{-}Sébastien Dessureault, Daniel Massicotte}%
\date{\today}%
\begin{document}%
\normalsize%
\maketitle%
\justify%
\section*{ABSTRACT}%
\label{sec:ABSTRACT}%
Clustering data is a popular feature in the field of unsupervised machine learning. Most algorithms aim to find the best method to extract consistent clusters of data, but very few of them intend to cluster data that share the same intersections between two features or more. This paper proposes a method to do so. The main idea of this novel method is to generate fuzzy clusters of data using a Fuzzy C-Means (\textit{FCM}) algorithm. The second part involves applying a filter that selects a range of minimum and maximum membership values, emphasizing the border data. A $\mu$ parameter defines the amplitude of this range.  
It finally applies a \textit{k-means} algorithm using the membership values generated by the \textit{FCM}. Naturally, the data having similar membership values will regroup in a new crispy cluster. The algorithm is also able to find the optimal number of clusters for the \textit{FCM} and the \textit{k-means} algorithm, according to the consistency of the clusters given by the Silhouette Index (SI). The result is a list of data and clusters that regroup data sharing the same intersection, intersecting two features or more. \textit{ck-means} allows extracting the very similar data that does not naturally fall in the same cluster but at the intersection of two clusters or more. The algorithm also always finds itself the optimal number of clusters. \newline

\noindent Keywords: \textit{ck-means}, \textit{FCM}, \textit{k-means}, fuzzy clustering, unsupervised learning, silhouette index

\begin{multicols}{2}%
\section{Introduction}%
\label{sec:Introduction}%
% Contexte, description du problème
Unsupervised learning is mainly used to solve clustering problems and dimensionality reduction. Regarding clustering, different methods are used to discriminate the data, regrouping it into clusters of similar elements. The problem is always the same: Find the best way to regroup similar data relative to their cluster. Different approaches compete to perform the most consistent clusters of data. This \textit{ck-means} method tackles the clustering problem from a different angle. It aims to cluster the data that shares the same intersection between features. In other words, \textit{ck-means} method can be used to extract the data in the border spaces of the features.  

% Techniques populaires
Let us see the most common algorithms used to cluster data. They all can be evaluated in terms of scalability, geometry, transductivity, and outliers management.

Affinity propagation \cite{dueck_non-metric_2007} can be interesting because the algorithm will choose the best number of clusters for a particular dataset on its own. Although, it is hard to scale with big datasets. MeanShift \cite{cheng_mean_1995} and Spectral clustering \cite{ng_spectral_2001} are also not very scalable, since they requires to find the nearest neighbours relative to some centroids.  Hierarchical clustering and Agglomerative Clustering \cite{fernandez_solving_2008} process not only the clusters but also multiple levels of subclusters. It can be visualized using dendrograms. DBScan \cite{tran_revised_2013} performs well when the clusters have different shapes but may have some difficulties when the cluster densities are different and are not very scalable. Optics \cite{ankerst_optics_1999} is similar to DBScan, but performs better when the density varies between clusters. It also requires minimal parameter tuning. Birch \cite{zhang_birch_1996} uses hierarchies and requires few resources. It is one of the best algorithms for large datasets.  

% \textit{k-means}
This novel method uses a \textit{k-means} algorithm along with a Silhouette Index (SI) to evaluate its performance, combined with a c-means algorithm (\textit{FCM}). From all the known clustering methods, \textit{k-means} algorithm \cite{hartigan_algorithm_1979} \cite{ahmed_k-means_2020} is the most popular.  Several declinations of this algorithm are proposed in the literature \cite{pham_selection_2005} \cite{likas_global_2003} \cite{krishna_genetic_1999}. It is also used in multiple applications \cite{alam_automatic_2019}. The \textit{k-means} algorithm is known to by a crispy algorithm, by opposition to a fuzzy algorithm based in the fuzzy logic principles. This crispy aspect of this \textit{k-means} method is one of its limitation. 

% Silhouette Index
Since it is a popular method, several metrics have been developed to evaluate the performance of the \textit{k-means} clustering process.  The Silhouette Index (SI) \cite{rousseeuw_silhouettes_1987} \cite{noauthor_towards_2011} \cite{starczewski_performance_2015} \cite{dudek_silhouette_2020} \cite{rawashdeh_crisp_2012} is widely used to evaluate the consistency of every data, in every cluster. It also supply an average for all the clusters.  The SI can be represented in a very intuitive way in a graphic. 

% C-Means
The strategy to emphasize the intersection data in this research is made with a \textit{FCM} algorithm \cite{bezdek_fcm_1984}\cite{askari_fuzzy_2021} \cite{arora_fuzzy_2019} . This fuzzy variation of a \textit{k-means} algorithm is also reputed to be helpful.  The advantage is this method is that each element are not included in only one cluster.  Each data may be owned, at a certain level (called "membership"), to the clusters. Like the \textit{k-means} algorithms, many variations exist to improve the original algorithm \cite{havens_fuzzy_2012} \cite{wang_global_2006} \cite{pal_possibilistic_2005} \cite{chuang_fuzzy_2006}.  
 
% Dataset description
The datasets used to validate this method has been generated using the \textit{Scikit-Learn} framework \cite{kramer_scikit-learn_2016} and \textit{datasets.make_classification()} \cite{noauthor_sklearndatasetsmake_classification_nodate} functions using different parameters. It was used to produce different datasets based on classification problems. When being called with the same parameters, the generated datasets are always the same, being fully reproducible.  

% Contribution
The main contribution of this paper is to propose a method that combines the \textit{FCM} and the \textit{k-means} algorithm, along with the SI, aiming to do clustering of intersection data. In other words, it regroups the data sharing the same borders. Also, unlike \textit{k-means} and the \textit{FCM} algorithms, \textit{CK-means} is able to find the best number of cluster, according to the best resulting consistency.  

This novel method might be helpful in a wide variety of domains: smart cities, health, manufacturers and sports statistics, to name a few. 

% Structure
The next sections of this paper are organized with the following structure: Section \ref{sec:Methodology} describes the proposed methodology. Section \ref{sec:Results} presents the results. Section \ref{sec:Discussions} discusses about the results and their meaning and Section \ref{sec:Conclusion} concludes this research.

\section{Methodology}%
\label{sec:Methodology}%

\subsection{Preprocessing of dataset}%
\label{subsec:Sub_methodology_preprocess}%
% Description plus complète du dataset.
Section \ref{sec:Introduction} mentions that the datasets are generated by  
the \textit{datasets.make_classification()} of the\textit{Scikit-learn} framework.  The parameters are used to select the different characteristics of the dataset, like the number of rows and features. Here is a definition of the \textit{datasets.make_classification()} parameters. \textit{n_sample}: The number of rows or generated data.  \textit{n_features}: The number of columns or features. \textit{n_classes}: The number of fields that targets the classification data. \textit{shuffle}: When equal to true, it randomly changes the order of the data.  \textit{random_state}: The seed is used to generate the data randomly. The same seed means the same generated data. This is important to have better reproducibility.

\subsection{Architecture}%
\label{subsec:Sub_methodology_architecture}%

% Description de l'architecture
The Figure \ref{fig:Archi} shows the architecture of the \textit{ck-means} method. 
\noindent\begin{center} \includegraphics[width=\columnwidth]{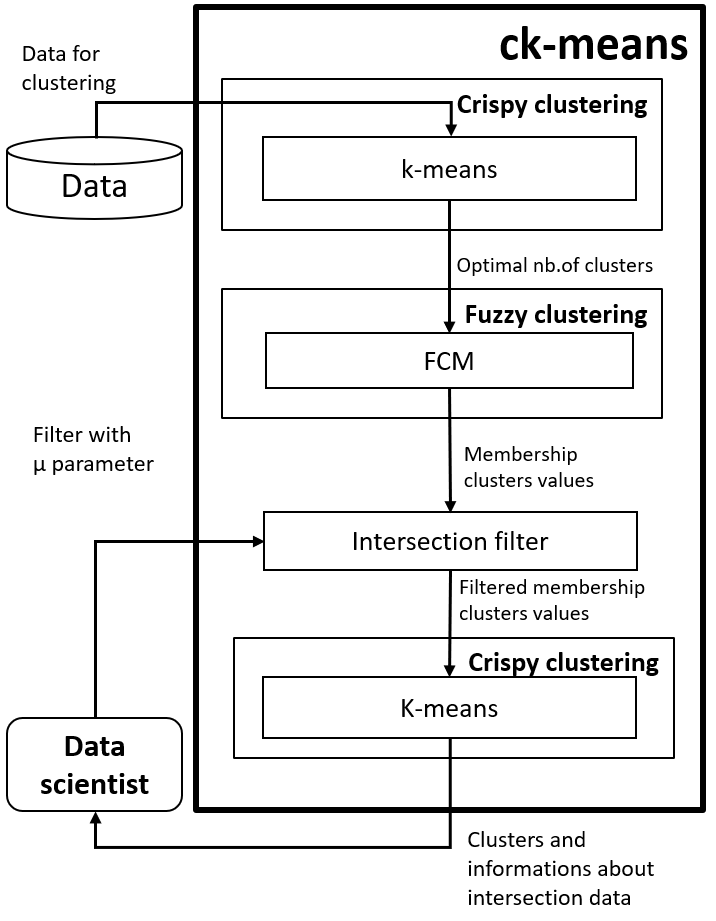}%
\captionof{figure}{Architecture of \textit{ck-means} method \label{fig:Archi}}
\end{center}%

This figure shows that the first part is a standard clustering using the k-mean algorithm. It finds the optimal number of clusters for the dataset. It is followed by a fuzzy clustering process using the \textit{FCM} algorithm. As in every \textit{FCM} process, the output is a list of clusters, where every data is identified to every cluster at a certain level called "membership". This output is the input of the next step, consisting of a filter that keeps only the data located in the intersections values in the dataset. The size of the intersection range is defined by a parameter called $\mu$. The filtered data are finally sent to a  \textit{k-means} clustering. The results are presented in the form of a list of values/clusters and some graphics.  

\subsection{FCM: c-mean algorithm}%
\label{subsec:Sub_FCM}%
Like the \textit{k-means} algorithm, \textit{FCM} (Fuzzy c-means) is a unsupervised learning algorithm to cluster data. The difference between those two techniques is that \textit{FCM} uses principles of fuzzy logic, while \textit{k-means} use traditional logic (also called "crispy logic" in opposition to "fuzzy").  The main difference is that a \textit{k-means} algorithm assign a cluster number to each data, while the \textit{FCM} assign a membership level for each data, for each cluster.  For instance, if the algorithm divided data into 3 clusters, each data will have 3 resulting values explaining the membership level (a normalized value between 0 and 1).  In this 3 clusters context, a data could have for instance membership values of 0.4, 0.1, and 0.5 for cluster 1, cluster 2 and cluster 3, respectively.  Equations (\ref{eq:FCM1}), (\ref{eq:FCM2}) and (\ref{eq:FCM3}) defines the \textit{FCM} algorithm.

% https://sites.google.com/site/dataclusteringalgorithms/fuzzy-c-means-clustering-algorithm

\begin{dmath}
\mu_{ij} = 1 / {\sum _{k=1}^{c} \left( d_{ij} / d_{ik} \right) ^{(2/m-1)}}   
\label{eq:FCM1}
\end{dmath}%

\begin{equation}
v_{j} =  \frac{( \sum _{i=1}^{n} (\mu_{ij})^{m} x_{i})}{( \sum _{i=1}^{n} (\mu_{ij})^{m})},  \forall j \in [1,c]
\label{eq:FCM2}
\end{equation}%
%\center$\forall j \in [1,c]$
%\justify

Where $n$ is the number of data elements, $m$ is the fuzziness index in domain $m \in [1, \infty]$, $\mu_{ij}$ is membership of the $i^{th}$ data to $j^{th}$ cluster centroid. $v_{j}$ is the $j^{th}$ cluster centroid.  $c$ is the number of centroids and $d_{ij}$ represents the Euclidean distance between $i^{th}$ data and $j^{th}$ cluster centroid.

\begin{equation}
J(U,V) =  \sum _{i=1}^{n}  \sum _{j=1}^{c} (\mu_{ij})^{m}\parallel x _{i} - v _{j}\parallel ^{2}
\label{eq:FCM3}
\end{equation}%

\textit{FCM} aims to minimize the value of (\ref{eq:FCM3}).\newline $\parallel x _{i} - v _{j}\parallel$ is the Euclidean distance between $i^{th}$ data and $j^{th}$ cluster centroid.

\subsection{\textit{k-means} algorithm and SI metric}%
\label{subsec:Sub_kmeans}%

This process aims to create some clusters using an unsupervised learning technique called \textit{k-means}. It is necessary to use an unsupervised learning technique since there is no label for each input data. The \textit{k-means} process will assign to each data a reference cluster according to the similarity level of their features. In this case, the input of this algorithm is not the raw data but the output coming from the filtered \textit{FCM} algorithm that was executed first. This \textit{FCM} output is the membership value of the data for each cluster. In other words, a membership level for each data, for each cluster.  

\noindent Equation (\ref{eq:kmeans}) defines the \textit{k}-means clustering equation where \textit{J} is a clustering function, \textit{k} is the number of clusters, \textit{n} is the number of features, $x_{i}^{(j)}$ is the input (feature \textit{i} in cluster \textit{j}) and $c_{j}$ is the centroid for cluster \textit{j}. To find the centroids values, the algorithm must try some random values and select the ones who minimize the inertia value. This inertia is a standard metric to evaluate the cluster consistency with \textit{k}-means.      

\begin{equation} \label{eq:kmeans} 
J=\sum _{j=1}^{k}\sum _{i=1}^{n}\left\| x_{i}^{(j)} -c_{j} \right\| ^{2}    
\end{equation} 

Along with inertia, there are several other metrics to measure clustering performance. Each metric is not compatible with every clustering algorithm. The SI metric is widely used to evaluate the consistency of clusters generated with \textit{k-means}. This SI metric is documented in the work of \cite{Rousseeuw_2009} and \cite{gueorguieva_mmfcm_2017}. Three equations define the SI metric. First, the distance between each point and the center of its cluster is defined by (\ref{eq:Silhouette1}). Then, the distance between the center of each cluster is shown in (\ref{eq:Silhouette2}). At last, (\ref{eq:Silhouette3}) uses the result of (\ref{eq:Silhouette1}) and (\ref{eq:Silhouette2}) to calculate the final SI score that indicates the quality (the consistency) of the clustering process.

For data point $i \in C_{I}$:
\begin{equation} \label{eq:Silhouette1} 
a(i)=\frac{1}{\left|c_{i} -1\right|} \sum _{j\in c_{i} i\ne j}d(i,j)  
\end{equation} 
$a(i)$ is the mean distance between \textit{i} and other data points in the same cluster.  The number of point in the cluster \textit{i} is $|C_{I}|$. 
$d(i,j)$ is the distance between data points \textit{i} and \textit{j} in the cluster $C_{I}$ 

Then, for data point $i \in C_{I}$:

\begin{equation} \label{eq:Silhouette2} 
b(i)={\mathop{\min }\limits_{k\ne i}} \frac{1}{\left|c_{k} \right|} \sum _{j\in c_{k} }d(i,j)  
\end{equation} 
$b(i)$ is the minimal mean distance of \textit{i} to all points in any other cluster, of which \textit{i} is not a member. This cluster having minimum mean dissimilarity is the "neighboring cluster" of \textit{i} because it is the next best fit cluster for point \textit{i}.  Having defined \textit{a(i)} and \textit{b(i)}, the final equation defines \textit{s(i)} which is the final silhouette index.  

\begin{equation} \label{eq:Silhouette3} 
s(i)=\frac{b(i)-a(i)}{\max \left(a(i),b(i)\right)} ,\; \; {\rm if}\; \left|C_{i} \right|>1 
\end{equation} 

SI domain of values is defined from -1 to +1. Values from -1 to 0 indicate a wrong classification. SI values from 0 to 1 indicate the points associated with a good cluster. The higher the value, the better the cluster consistency \cite{Rousseeuw_2009}. 

\subsection{Intersection filter}%
\label{subsec:Sub_filter}%
The "intersection filter" goal is to remove data identified in only one cluster, having a membership too low to be considered in the intersection. For instance, data with membership values of 0.97, 0.03, or 0.0 would be removed since this data would not be considered at an intersection. The filter keeps data within a threshold defined by the user-defined $\mu$ parameter. This $\mu$ parameter defines the minimum and the maximum values of the thresholds when \textit{\textit{ck-means}} algorithm is called. Using $\mu$, the minimum value kept by the filter will be $0.5 - (\mu / 2)$ and the maximum will be $0.5 + (\mu / 2)$. For instance, having $\mu = 0.4$, the minimum would be 0.3, and the maximum would be 0.7.\newline 

Hence, the data scientist can narrow or enlarge the amount of data included in the intersection.

\section{Results}%
\label{sec:Results}%

The following cases test some different aspects of this novel method. For each case, different parameters and a different dataset. For each dataset, "Nb.samples" is the number of rows in the dataset.  \textit{Nb.features} is the number of different features and \textit{Nb.Centroids} is the number of natural clusters generated. The \textit{Cluster standard deviation} is a metric of consistency. The higher this value, the less the cluster will be consistent, and the points might be mixed up with others points of other clusters. Finally, the \textit{Seed} generates the data randomly. Having the same seed means that the same data sequence will be generated each time. This is very useful for reproducibility purposes. 

\subsection{Case 1}%
\label{subsec:Case1}%

% Text describing the utility of this case
This first case is a basic example of the intersection extraction using only 2 clusters and a 2 features dataset. A 2 features dataset can be easily presented using 2d graphics. Table \ref{table:table_datasets1} shows the dataset used for this first case. \newline

\begin{minipage}[htb]{1.0\columnwidth}
\captionof{table}{Datasets for case 1, using $\mu$=0.4 }% 
\begin{tabular}{lr} \hline 
\label{table:table_datasets1}%
\textbf{Dataset} & \textbf{Value}  \\
\hline
Nb.samples & 500  \\
Nb.features & 2  \\
Nb.centroids & 2  \\
Clusters standard deviation & 5.0 \\
Seed & 15   \\
\hline
\end{tabular}
\end{minipage}\newline

% Important figures and explanations
Fig. \ref{fig:c1_dist} shows the data distribution. 2 clusters have been found. The red points represent the intersection data extracted with the \textit{FCM} algorithm and have a membership in the $\mu=0.4$ range (between 0.3 and 0.7). 

\noindent\begin{center} \includegraphics[width=\columnwidth]{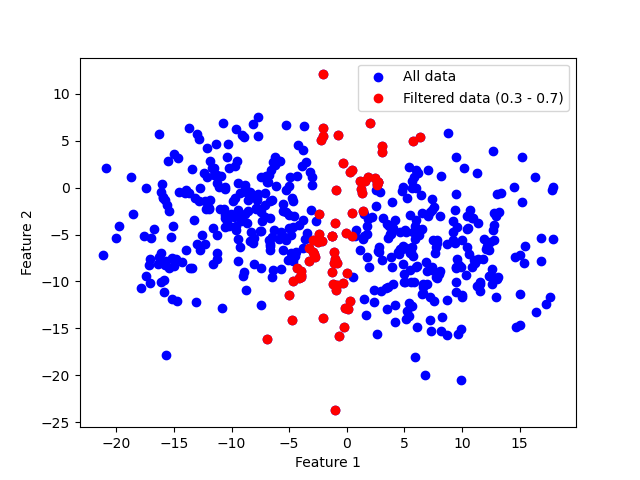}
\captionof{figure}{Distribution of the data, emphasizing the intersection data (in red, using $\mu=0.4$) \label{fig:c1_dist}}
\end{center}%

The same intersection data are filtered ($\mu=0.4$) and magnified in Fig. \ref{fig:c1_cmeans}. The colour intensity represents the level of membership in their respective cluster. 

\noindent\begin{center} \includegraphics[width=\columnwidth]{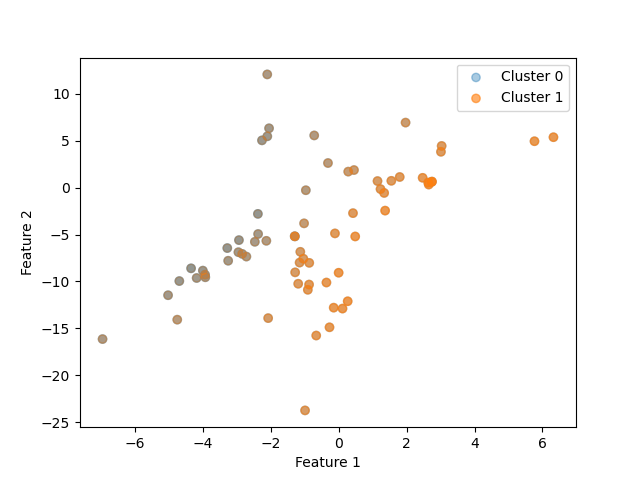}
\captionof{figure}{Intersection data magnified and colored according to their cluster membership level. \label{fig:c1_cmeans}}
\end{center}%

In Fig. \ref{fig:c1_membership}, we can see the distribution of the membership value that has been produced by the \textit{FCM} algorithm. Since 2 clusters have been found, the result is linear because the sum of the 2 memberships is always 1. If the first membership value is 0.2, the other value will automatically be 0.8. 

\noindent\begin{center} \includegraphics[width=\columnwidth]{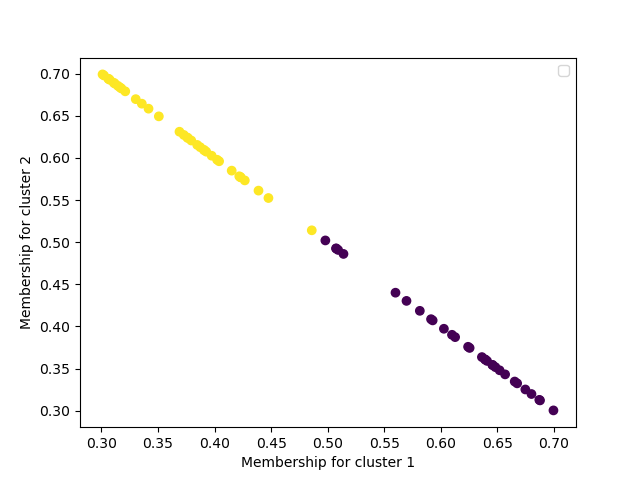}
\captionof{figure}{Distribution of the membership of the data in the intersection. \label{fig:c1_membership}}
\end{center}%

The final step is to cluster the filtered resulting membership using a \textit{k-means} algorithm, as shown in Fig. \ref{fig:c1_kmeans}. This process may seem useless in these 2 clusters' context since data are already quite regrouped. Although, the clusters will be distinct when the number of clusters is in higher dimensions. Also, even with only two clusters, there might be some natural clusters in the intersection that was not obvious before the filtering process. 

\noindent\begin{center} \includegraphics[width=\columnwidth]{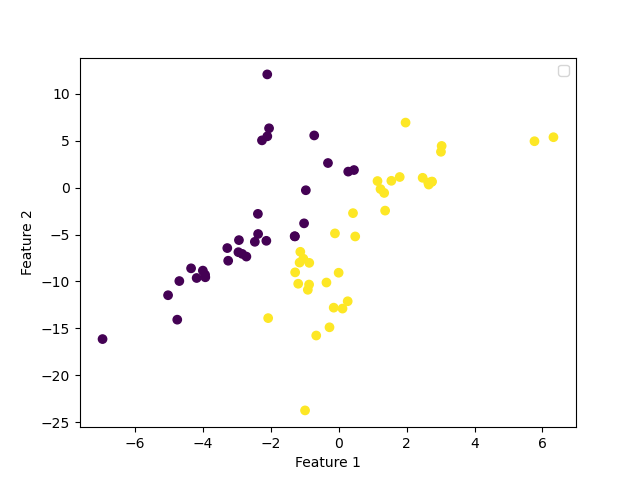}
\captionof{figure}{Final \textit{k-means} clustering of the intersection data \label{fig:c1_kmeans}}
\end{center}%
 
The last graphic (Fig. \ref{fig:c1_Silhouete}) is the proof of the consistency of the intersection clusters. It shows a SI of 0.7199, which means a very consistent clustering. The grey and green bands are the clusters composed of the data. X axe is the value of the clustering consistency (a normalized distance from each point to their cluster centroid). We can see no misplaced value (negative values on X axe). The red dotted line defines the SI value of the whole clustering process. 

\noindent\begin{center} \includegraphics[width=\columnwidth]{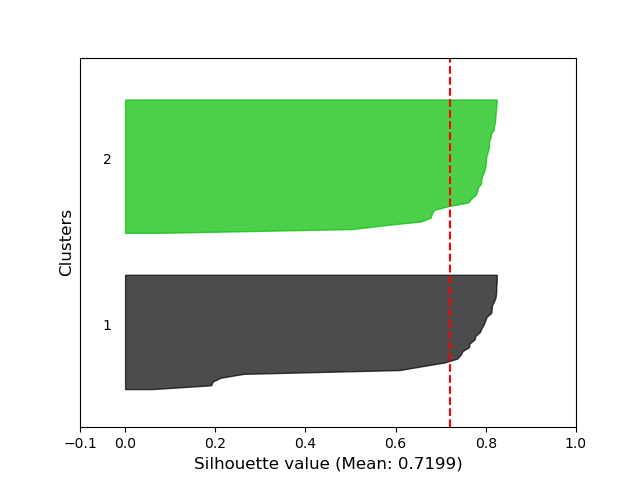}
\captionof{figure}{Consistency proof of the intersection data using SI, after a \textit{k-means} clustering. \label{fig:c1_Silhouete}}
\end{center}%

% conclusion of the case
This first basic case showed a good consistency of the extracted intersection data using a \textit{FCM} algorithm and clustered using a \textit{k-means} algorithm.

\subsection{Case 2}%
\label{subsec:Case2}%

% Text describing the utility of this case
This second example shows the frequent case of no data in the intersection. In this case, the utility of the \textit{ck-means} algorithm for the data scientist is to learn that there is no intersection data corresponding to a certain filter range (here: $\mu=0.4$).   Table \ref{table:table_datasets2} presents the used dataset. \newline

\begin{minipage}[htb]{1.0\columnwidth}
\captionof{table}{Datasets for case 2, using $\mu$=0.4 }% 
\begin{tabular}{lr} \hline 
\label{table:table_datasets2}%
\textbf{Dataset} & \textbf{Value}  \\
\hline
Nb.samples & 500  \\
Nb.features & 2  \\
Nb.centroids & 3  \\
Clusters standard deviation & 0.4 \\
Seed & 12   \\
\hline
\end{tabular}
\end{minipage}\newline

% Important figures and explanations
The Fig. \ref{fig:c2_dist} shows 3 very consistent clusters.  There is no intersection data. It would have been the red points. In this case, the \textit{ck-means} algorithm returns this graphic and a message saying, "No data are found in the intersection corresponding to the $\mu$ parameter.".

\noindent\begin{center} \includegraphics[width=\columnwidth]{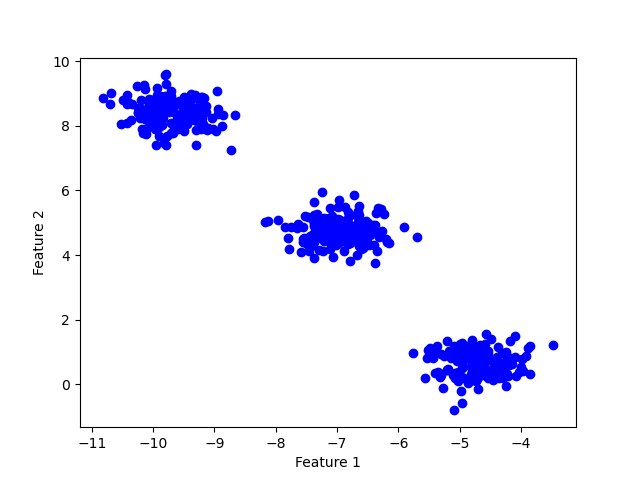}
\captionof{figure}{Distribution of the data when there are no intersecting data. \label{fig:c2_dist}}
\end{center}%

% conclusion of the case
This case showed that the \textit{ck-means} algorithm behaves correctly when no data is located at the clusters' intersections.

\subsection{Case 3}%
\label{subsec:Case3}%
% Text describing the utility of this case
This case is similar to case 1 but adds a feature to obtain 3 dimensions data. There are also only 2 centroids, suggesting that the \textit{FCM} algorithm will work with 2 clusters.   The dataset used for is shown in Table \ref{table:table_datasets3}. \newline

\begin{minipage}[htb]{1.0\columnwidth}
\captionof{table}{Datasets for case 3, using $\mu$=0.4 }% 
\begin{tabular}{lr} \hline 
\label{table:table_datasets3}%
\textbf{Dataset} & \textbf{Value}  \\
\hline
Nb.samples & 500  \\
Nb.features & 3  \\
Nb.centroids & 2  \\
Clusters standard deviation & 3.5 \\
Seed & 15   \\
\hline
\end{tabular}
\end{minipage}\newline

% Important figures and explanations
The Fig. \ref{fig:c3_dist} presents the distribution of the data. As foreseen by the number of centroids, 2 clusters have been found. The orange points represent the intersection data extracted with the \textit{FCM} algorithm and a filter of $\mu=0.4$.

\vfill\null
\columnbreak 

\noindent\begin{center} \includegraphics[width=\columnwidth]{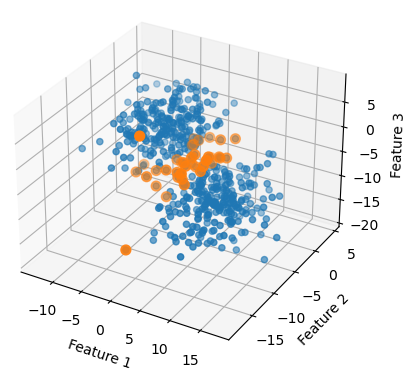}
\captionof{figure}{Distribution of the data, emphasizing the intersection data (in orange), using $\mu=0.4$ \label{fig:c3_dist}}
\end{center}%

Fig. \ref{fig:c3_membership} shows the distribution of the membership value that has been produced by the \textit{FCM} algorithm. As in case 1, the result is linear because the sum of the 2 memberships is always 1. Although, in this case, the \textit{k-means} algorithm found that those membership values are more consistently clustered in 3 groups, according to the SI of 0.7057. The result of this process is shown in Fig. \ref{fig:c3_kmeans}.    

\noindent\begin{center} \includegraphics[width=\columnwidth]{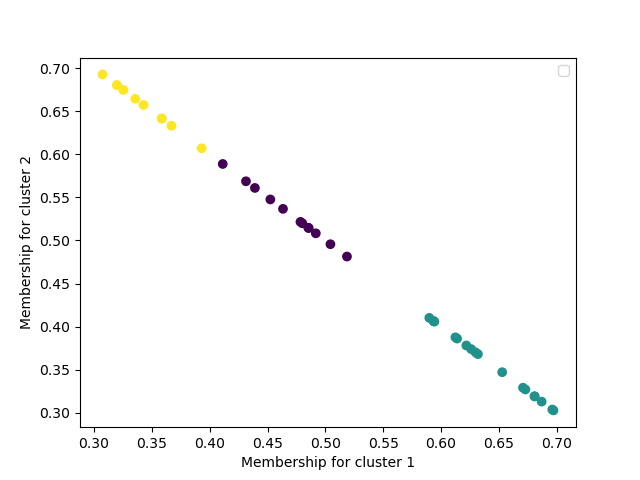}
\captionof{figure}{Distribution of the membership of the data in the intersection. \label{fig:c3_membership}}
\end{center}%

\vfill\null
\columnbreak 

\noindent\begin{center} \includegraphics[width=\columnwidth]{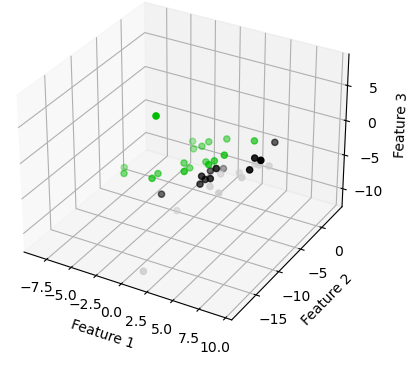}
\captionof{figure}{Final \textit{k-means} clustering of the intersection data \label{fig:c3_kmeans}}
\end{center}%
 
% conclusion of the case
This case tests the \textit{ck-means} algorithm using 3 dimensions data.  It proves that the intersection data are correctly extracted, filtered, and clustered.

\subsection{Case 4}%
\label{subsec:Case4}%

% Text describing the utility of this case
This 4th case aims to extract intersection data when there are multiple centroids in the data. The $\mu$ parameter has been increased to 0.6 for this case, meaning a membership function filtered between 0.3 to 0.7. The final \textit{k-means} clustering process will be particularly significative in this case since the data are located in distinct parts of the graphic. Table \ref{table:table_datasets4} presents the parameter used to generate the dataset. \newline

\begin{minipage}[htb]{1.0\columnwidth}
\captionof{table}{Datasets for case 4, using $\mu$=0.6 }% 
\begin{tabular}{lr} \hline 
\label{table:table_datasets4}%
\textbf{Dataset} & \textbf{Value}  \\
\hline
Nb.samples & 500  \\
Nb.features & 2  \\
Nb.centroids & 5  \\
Clusters standard deviation & 1.0 \\
Seed & 2   \\
\hline
\end{tabular}
\end{minipage}\newline

% Important figures and explanations
Fig. \ref{fig:c4_dist} displays the data distribution. Although there were 5 centroids in the data, 4 clusters have been found due to a significant standard deviation and 2 close centroids. The red points represent the intersection data using parameter $\mu=0.6$. Those filtered data are shown in Fig. \ref{fig:c4_cmeans}. The colour intensity explains the membership level of their cluster. 

\noindent\begin{center} \includegraphics[width=\columnwidth]{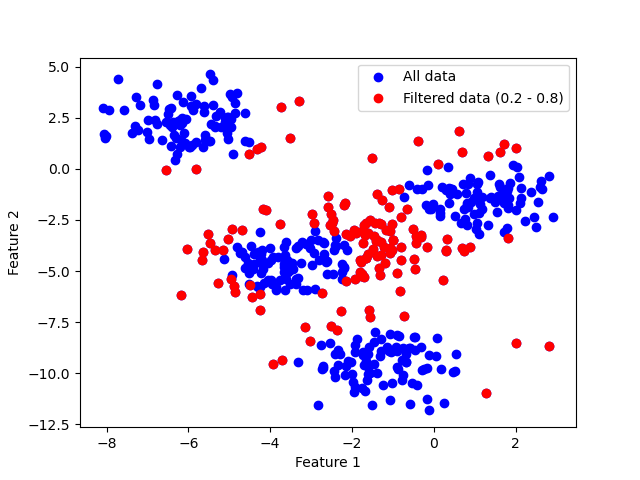}
\captionof{figure}{Distribution of the data, emphasizing the intersection data (in red, using $\mu=0.4$) \label{fig:c4_dist}}
\end{center}%

\noindent\begin{center} \includegraphics[width=\columnwidth]{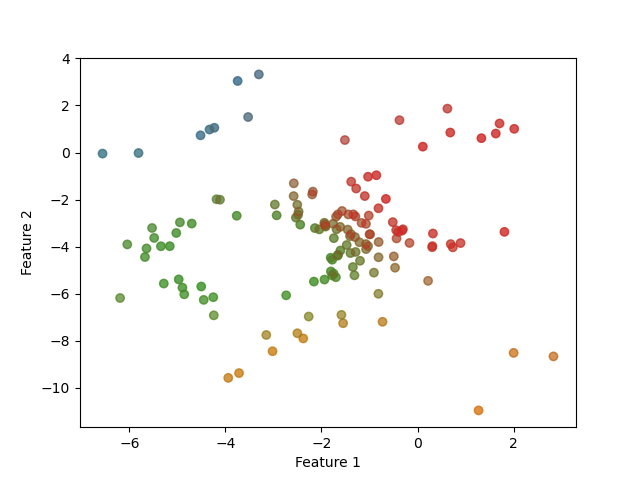}
\captionof{figure}{Intersection data magnified and colored according to their cluster membership level. \label{fig:c4_cmeans}}
\end{center}%

As in the other cases, a \textit{k-means} algorithm (Fig. \ref{fig:c4_kmeans}) processed the data to rebuild new clusters from the intersection data. In this case, 4 clusters was found to maximize a SI of 0.6492 (Fig. \ref{fig:c4_Silhouete}). 

\vfill\null
\columnbreak  
\pagebreak

\noindent\begin{center} \includegraphics[width=\columnwidth]{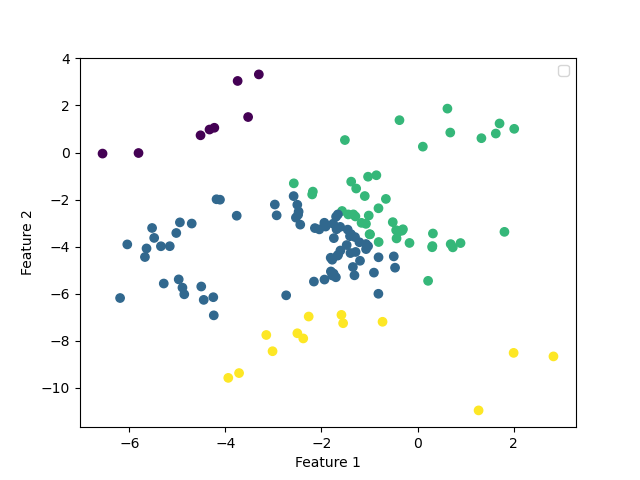}
\captionof{figure}{Final \textit{k-means} clustering of the intersection data \label{fig:c4_kmeans}}
\end{center}%
 
\noindent\begin{center} \includegraphics[width=\columnwidth]{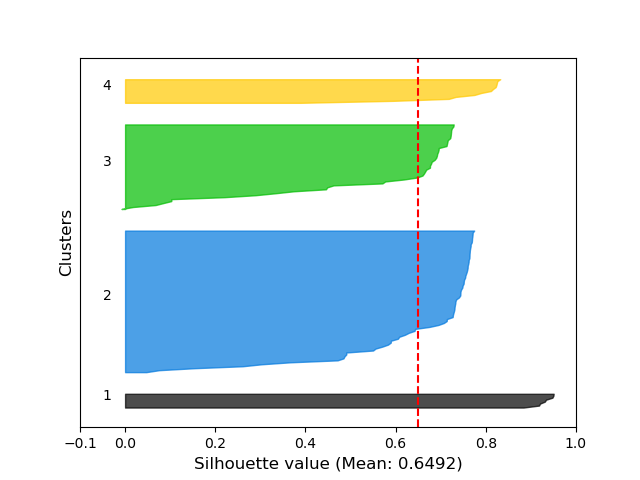}
\captionof{figure}{Consistency proof of the intersection data using SI, after a \textit{k-means} clustering. \label{fig:c4_Silhouete}}
\end{center}%

% conclusion of the case
Even with this higher number of clusters, the \textit{ck-means} algorithm succeeds in extracting the intersection data and in clustering the results based on the membership values generated by the \textit{FCM} algorithm.

\subsection{Case 5}%
\label{subsec:Case5}%

% Text describing the utility of this case
This case uses higher dimensions of features (6), so there is graphically a challenge to illustrate the validity of the process and the final clustering consistency. Parameters used to generate the dataset are shown in Table \ref{table:table_datasets5}. \newline 

\vfill\null
\columnbreak 

\begin{minipage}[htb]{1.0\columnwidth}
\captionof{table}{Datasets for case 5, using $\mu$=0.6 }% 
\begin{tabular}{lr} \hline 
\label{table:table_datasets5}%
\textbf{Dataset} & \textbf{Value}  \\
\hline
Nb.samples & 500  \\
Nb.features & 6  \\
Nb.centroids & 5  \\
Clusters standard deviation & 1.5 \\
Seed & 1   \\
\hline
\end{tabular}
\end{minipage}\newline

% Important figures and explanations
The first \textit{FCM} algorithm generates membership data that can be displayed in a 3d space (Fig. \ref{fig:c5_Membership}).  It clearly shows a future \textit{k-means} process of 3 distincts clusters.    

\noindent\begin{center} \includegraphics[width=\columnwidth]{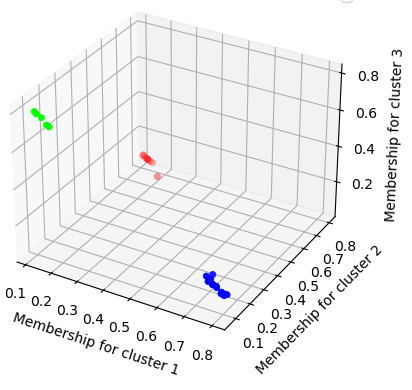}
\captionof{figure}{Membership values of the initial \textit{FCM} 3 clusters. \label{fig:c5_Membership}}
\end{center}%

To illustrate the validity of the \textit{ck-means} algorithm, Fig. \ref{fig:c5_Radar_A}, \ref{fig:c5_Radar_B} and \ref{fig:c5_Radar_C} present the 3 generated clusters resulting from the \textit{FCM} and the \textit{k-means} algorithm.  Those are stacked radar graphics of the clustered and filtered intersection data. The data has been normalized using a MinMax function to fit on the same scale radar graphic. 

\vfill\null
\columnbreak 

\noindent\begin{center} \includegraphics[width=7cm]{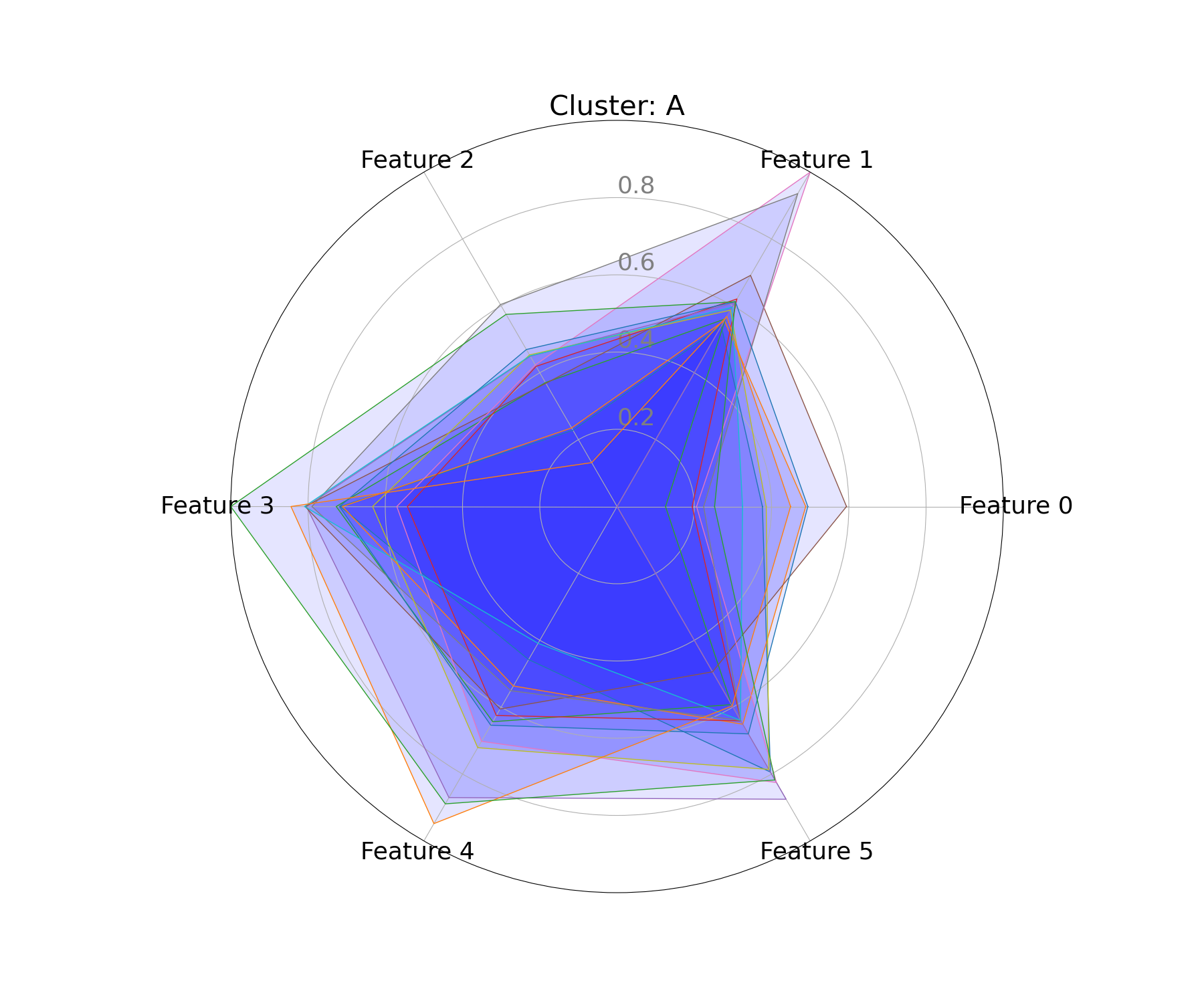}
\captionof{figure}{MinMax normalized stacked radar graphic presenting cluster A \label{fig:c5_Radar_A}}
\end{center}%

\noindent\begin{center} \includegraphics[width=7cm]{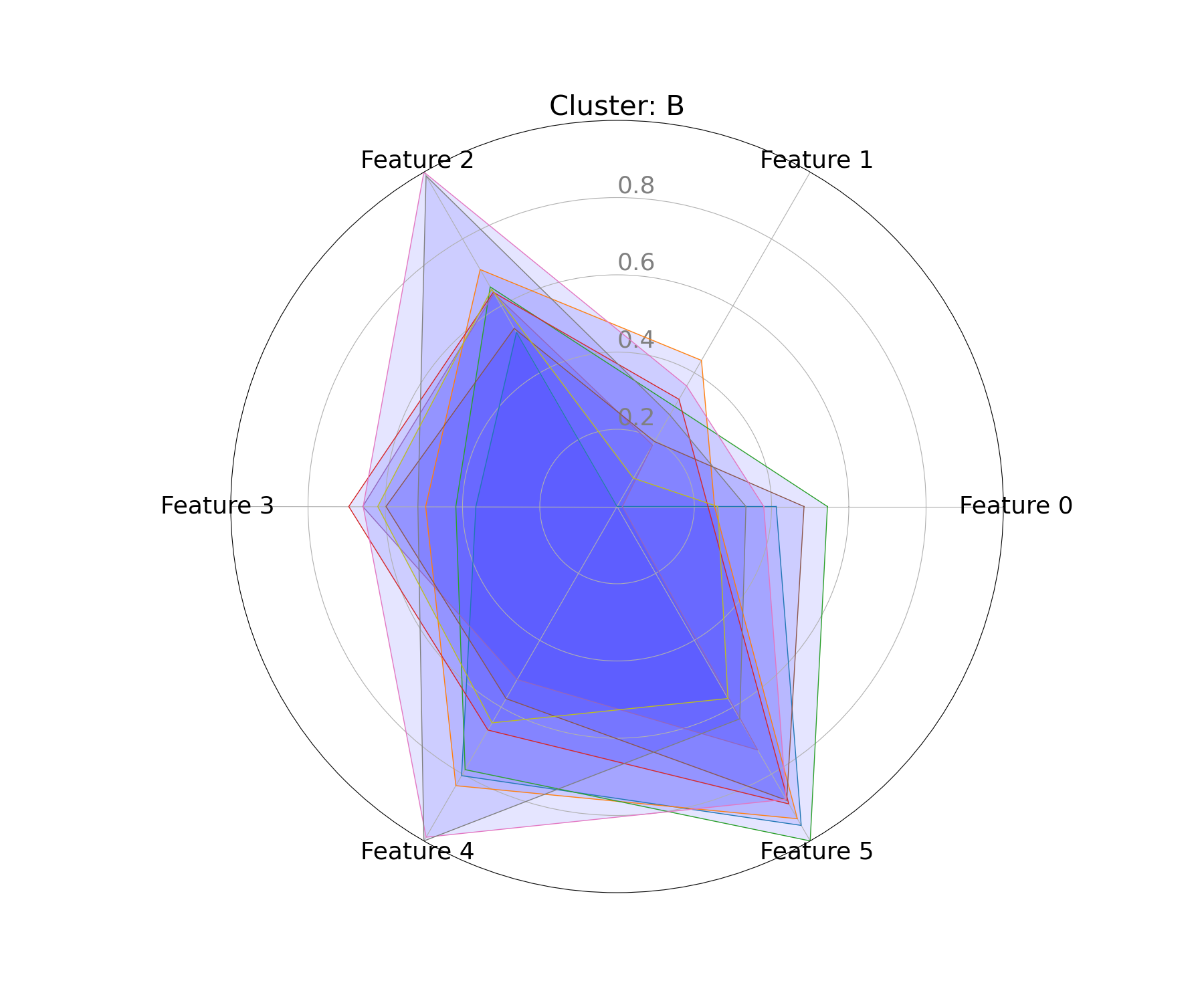}
\captionof{figure}{MinMax normalized stacked radar graphic presenting cluster B \label{fig:c5_Radar_B}}
\end{center}%

\noindent\begin{center} \includegraphics[width=7cm]{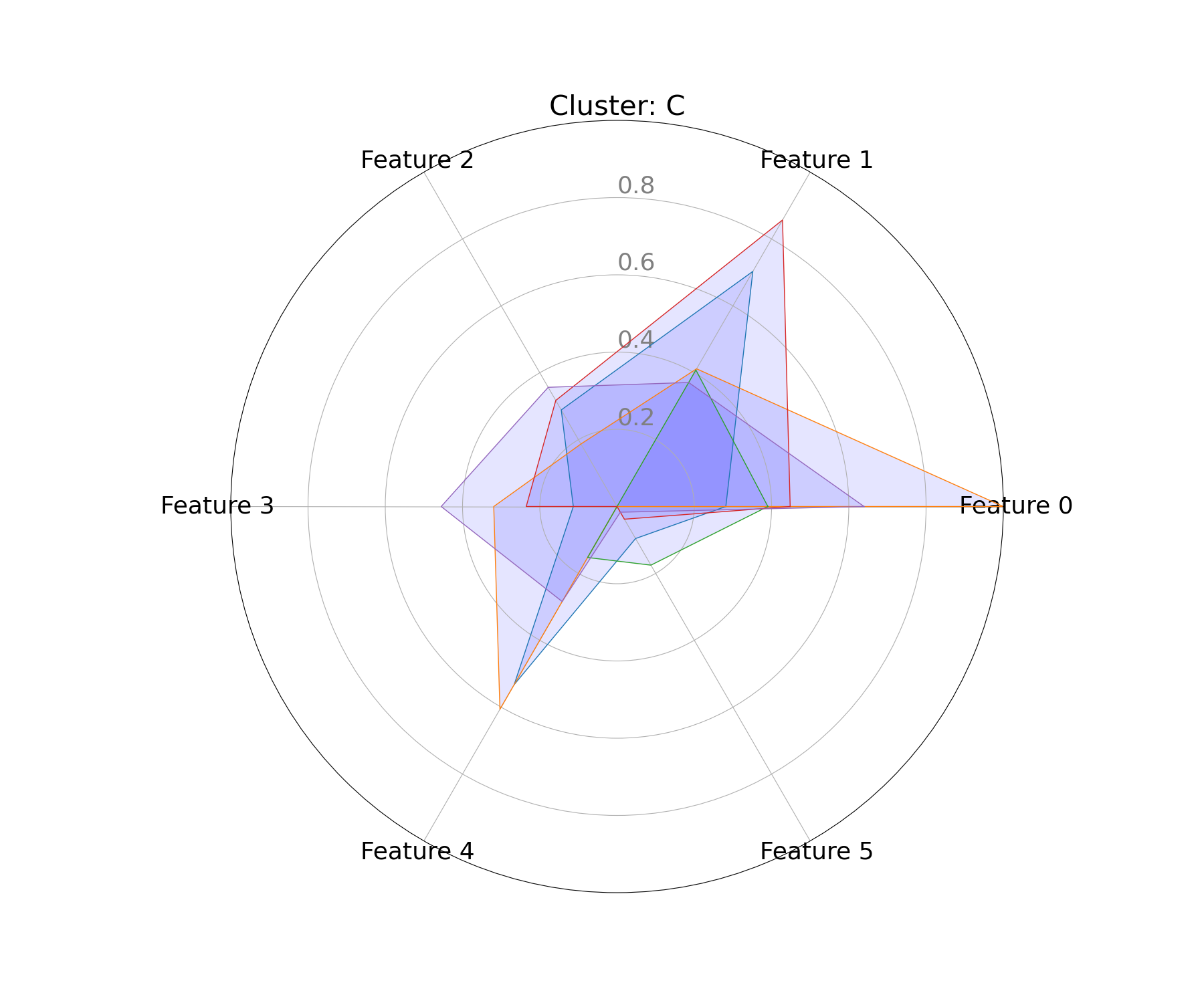}
\captionof{figure}{MinMax normalized stacked radar graphic presenting cluster C \label{fig:c5_Radar_C}}
\end{center}%

The consistency of the clustering process is very high, having a SI of 0.9374. 

% conclusion of the case
Although there was a challenge in visually presenting the validity of the process, this case showed that the \textit{ck-means} works well in a higher dimension context. 

\subsection{Case 6}%
\label{subsec:Case6}%
% Text describing the utility of this case
This final case tests the scalability of the \textit{ck-means} algorithm. The dimensionality of the samples (100,000) and of the features (30) is significantly higher as presented in Table \ref{table:table_datasets6}. Also, the filter parameter will keep most of the data using a $\mu$ value of 0.8. This case aims to see if, at this scale, the algorithm still works in a reasonable time. 

\begin{minipage}[htb]{1.0\columnwidth}
\captionof{table}{Datasets for case 1, using $\mu$=0.8 }% 
\begin{tabular}{lr} \hline 
\label{table:table_datasets6}%
\textbf{Dataset} & \textbf{Value}  \\
\hline
Nb.samples & 100000  \\
Nb.features & 30  \\
Nb.centroids & 5  \\
Clusters standard deviation & 1.5 \\
Seed & 1   \\
\hline
\end{tabular}
\end{minipage}\newline

% Important figures and explanations
The whole process was executed within minutes (5:36). The SI index was excellent at 0.9840. To illustrate the complexity of the final clustering process, Fig. \ref{fig:c6_Radar_A} shows an example of one cluster (Cluster A). 

\noindent\begin{center} \includegraphics[width=\columnwidth]{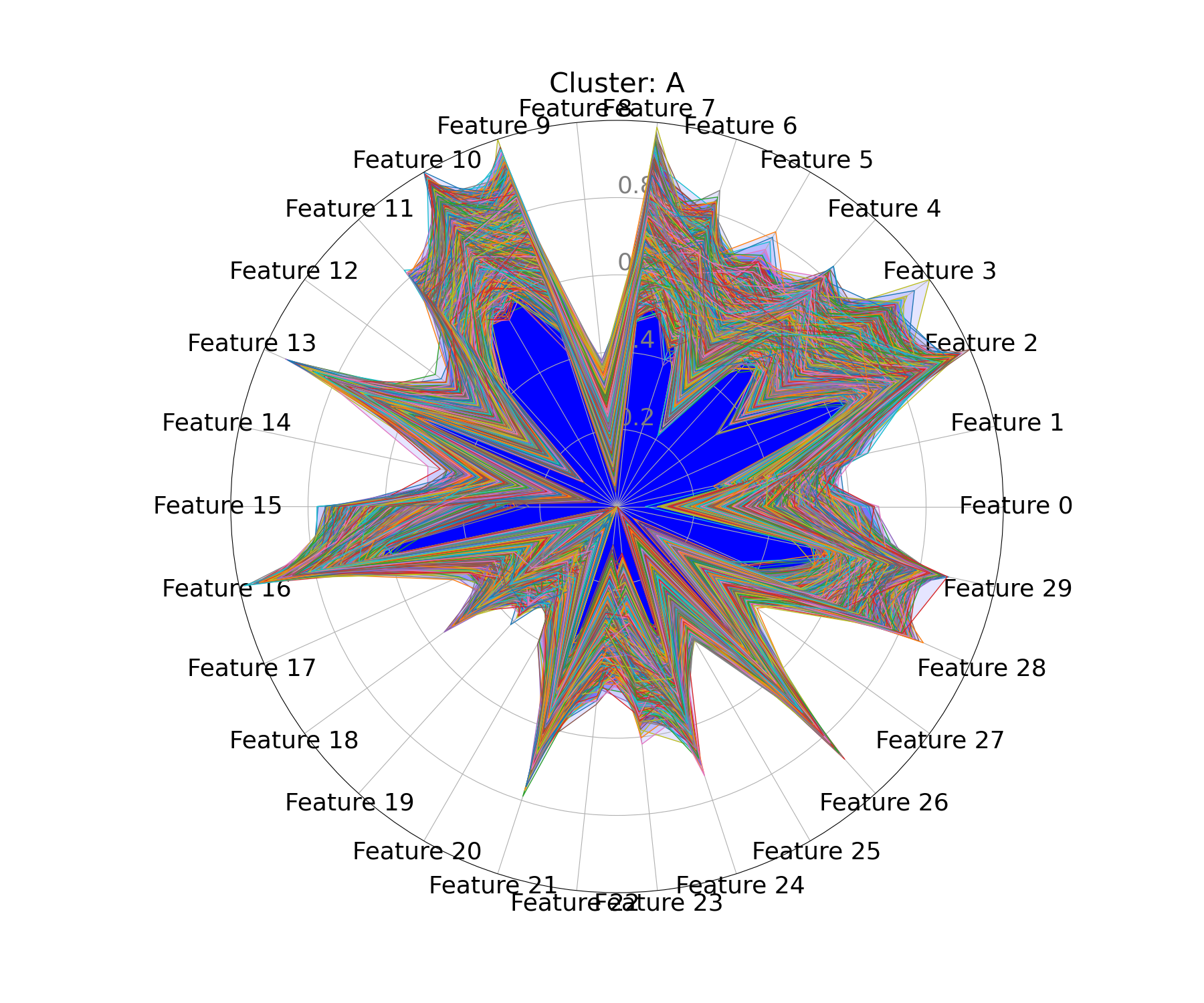}
\captionof{figure}{MinMax normalized stacked radar graphic  presenting cluster A \label{fig:c6_Radar_A}}
\end{center}%

% conclusion of the case
Even using a significantly larger dataset, the \textit{ck-means} passed the scalability test. The reason is that the algorithm does not add an extra embedded loop for the clustering process. There are only two embedded loops in the filtering process. This algorithm may scale without difficulty. 

\subsection{Comparison between this novel method and original methods}%
\label{subsec:Subcomparison}%
There are already multiple methods to extract clusters from datasets. In the introduction, several are referenced like Affinity propagation \cite{dueck_non-metric_2007}, MeanShift \cite{cheng_mean_1995}, Spectral clustering \cite{ng_spectral_2001}. There are also hierarchical clustering methods, like Agglomerative Clustering \cite{fernandez_solving_2008} and Birch  \cite{zhang_birch_1996} that can process multiple levels of subclusters, being visualized using dendrograms. DBScan \cite{tran_revised_2013} can handle the clusters having different shapes.  Finally the very known \textit{k-means} algorithm \cite{hartigan_algorithm_1979} and its fuzzy equivalent: the \textit{FCM} algorithm \cite{bezdek_fcm_1984}. Those are useful methods when it is time to extract data in clusters. Using those methods, the problem is that they can not extract specific data at the intersection of two clusters or more. In some situations, it is useful to address this specific data.  

To compare those methods with this novel \textit{ck-means}, we must refer to the mathematics field called the "set theory". While existing methods aim to determine data included in a set, \textit{ck-means} aims to determine data included in the intersection of two or more sets. 

Also, at the opposite of the \textit{k-means} and the \textit{FCM} algorithms, \textit{CK-means} can find alone the best number of clusters according to the best resulting SI metric. 

Using \textit{ck-means}, the metric to evaluate the resulting cluster consistency is the known SI metric. The clusters can be evaluated the same way as with a \textit{k-means} algorithm, for instance.  

Table \ref{table:compare} compares a sample of 20 data the \textit{k-means}, the \textit{FCM} and the \textit{ck-means} methods.\newline

\vfill\null
\columnbreak 

\noindent\begin{minipage}[htb]{1.0\columnwidth}
\captionof{table}{Comparison of the results for the \textit{k-means}, the \textit{FCM} and the \textit{ck-means} methods, using dataset defined in \ref{subsec:Case1}.
}% 
\begin{tabular}{rrrrrr} \hline 
\label{table:compare}%
\textbf{Data}   &  & \textbf{FCM}  &   & \textbf{k-m.}  & \textbf{ck-m.}  \\
\textbf{F.1} & \textbf{F.2} & \textbf{C.1}  & \textbf{C.2}  & \textbf{C.}  & \textbf{C.}  \\
\hline
2.79  & -6.84  & 0.90  & 0.09  & 1  & N.A.  \\
9.88  & -2.05  & 0.76  & 0.23  & 1  & N.A.  \\
2.49  & -2.00  & 0.77  & 0.22  & 1  & N.A.  \\
-6.67 & -6.68  & 0.10  & 0.89  & 0  & N.A.  \\
1.23  & -6.69  & 0.94  & 0.05  & 1  & N.A.  \\
-7.48 & 1.16   & 0.06  & 0.93  & 0  & N.A.  \\
-1.28 & 5.21   & 0.12  & 0.87  & 0  & N.A.  \\
9.50  & -1.00  & 0.95  & 0.04  & 1  & N.A.  \\
9.76  & -8.29  & 0.97  & 0.02  & 1  & N.A.  \\
1.05  & -4.29  & 0.95  & 0.04  & 1  & N.A.  \\
7.41  & -5.13  & 0.71  & 0.28  & 1  & N.A.  \\
5.48  & -2.88  & 0.93  & 0.06  & 1  & N.A.  \\
1.65  & -1.35  & 0.75  & 0.24  & 1  & N.A.  \\
-3.80 & 4.00   & 0.24  & 0.75  & 1  & N.A.  \\
-1.07 & 6.89   & 0.16  & 0.83  & 0  & N.A.  \\
...   & ...    & ...   & ...   & ...& ...   \\
1.79  & 1.13   & 0.61  & 0.38  & 1  & 1  \\  % 26
-5.03 & -11.47 & 0.35  & 0.64  & 1  & 0  \\  % 39
-4.70 & -9.96  & 0.32   & 0.67 & 1  & 0  \\  % 46
-4.00 & -8.84  & 0.33  & 0.66  & 1  & 0  \\  % 53
-2.26 & 5.04   & 0.33  & 0.66  & 1  & 0  \\  % 54

\hline
\end{tabular}
\end{minipage}\newline

Feature 1 (F.1) and Feature 2 (F.2) in this table identify the data. The result of the FCM clustering is presented as a membership value in cluster 1 (C.1) and cluster 2 (C.2). The cluster number associated with the \textit{k-means} (k-m.) and the \textit{ck-means} ck-m algorithms are also presented to compare the output of each one. Looking at the \textit{FCM} values, it can be seen that the \textit{ck-means} results are defined only for the values lying inside the range for the filter $\mu=0.4$. Hence, values between 0.3 and 0.7 are kept by the filter and clustered again using a \textit{k-means} method.
Note that all the values are trunked at 2 decimals. 

This table shows how the 3 methods are complementary.  \textit{k-means} is used for crispy clustering, \textit{FCM} is used for fuzzy clustering, and \textit{ck-means} uses both \textit{k-means} and \textit{FCM} to cluster data at the intersection of the clusters. 

\section{Discussions}%
\label{sec:Discussions}%
A novel unsupervised machine learning method to extract intersection data between clusters has been developed in this paper. The method has been validated using  different datasets generated with the \textit{make_classification()} function available in the \textit{scikit-learn} framework.  The parameters used to generate the data differed for each of the 6 use cases. The \textit{seed} parameter made the datasets fully reproducible. 

Note that every figure in this paper was generated by the \textit{ck-means} algorithm, making this method more explainable. 

%Case 1
The first case is the basic one. Fig. \ref{fig:c1_dist} shows the data distribution, emphasizing the filtered data (using $\mu$=0.4) located at the intersection of the 2 clusters. Fig. \ref{fig:c1_cmeans} shows the filtered data after this first clustering process using \textit{FCM} algorithm, using colour tones according to their membership level to the clusters. Fig. \ref{fig:c1_membership} is shown to explain the basic data (the membership level) used to do the next clustering (\textit{k-means}). The result of the \textit{k-means} clustering is shown in Fig. \ref{fig:c1_kmeans}. The consistency of this final clustering process is displayed in Fig. \ref{fig:c1_Silhouete}, showing a good cluster consistency (SI=0.7199).   This first basic example displays all the available graphics for the whole process.

%Case 2
Case 2 proves that the algorithm still works without intersection data. This can happen due to the structure of the data or a too-small $\mu$ parameter. Fig. \ref{fig:c2_dist} shows the distribution of the data.  Since there is no red dot in this graphic, we can conclude that the filter has eliminated all the data. This case may happen often. There is nothing to do if the data is very regrouped around the centroids. Nonetheless, the $\mu$ parameters may be augmented to a higher value, preventing the filter from dropping too much data.

%Case 3, 4, 5
In the 3rd case, 3-dimensional data show that the algorithm can also work in multiple dimensions. 3d graphics le Fig. \ref{fig:c3_dist} and \ref{fig:c3_kmeans} are used to represent the data. Case 4 processes higher cardinality in terms of number of centroids (5), showing the utility to cluster one more time the resulting data using a \textit{k-means} algorithm, as shown in \ref{fig:c4_dist} \ref{fig:c4_cmeans} \ref{fig:c4_kmeans}.  Once again, a SI of 0.6492 proves the result is consistent (\ref{fig:c4_Silhouete}). Case 5 aims to show the possibility of representing the data of 3d and higher in multiple clusters. There are 6 features distributed around 5 centroids for this case. There is a challenge in representing the data. It has been made using some stacked radar graphics, after the feature has been normalized using a MinMax function (Fig. \ref{fig:c5_Radar_A}, \ref{fig:c5_Radar_B} and \ref{fig:c5_Radar_C}). The very high value of the SI (0.9374) is explained clearly by Fig. \ref{fig:c5_Membership}. 

%Case 6
The last case (case 6) is the one that proves that the \textit{ck-means} algorithm can scale at a certain point. It uses 100000 samples and 30 features. It means that the intersection data were very isolated, without overlapping. It took less than 6 minutes to execute, showing that there are no combinatorial explosions. The result was good, showing a very high SI of 0.9840. It is also challenging to graphically represent a cluster consistency with 30 features. Nevertheless, Fig. \ref{fig:c6_Radar_A} succeeds in doing so, using a complex graphical structure, but showing an evident result. Without any doubt, this figure shows a good consistency for this cluster A.      

%Useful real-life cases.
Myriads of applications use this \textit{ck-means} algorithms. For instance, it can emphasize the district limits in a smart city context. It can be used with geographical coordinates to extract data at the borders and frontiers. It can be used in sports statistics. Some players may be very similar, but they are never regrouped together. For baseball statistics, the classical clustering method would not necessarily regroup players sharing a similar home run/stolen base/walk ratio. \textit{k-means} allows extracting those similar data that do not naturally fall in the same cluster. It is easy to foresee other applications for this novel algorithm. The domains of marketing, health and agriculture would be attractive, to name a few. 

\section{Conclusion}%
\label{sec:Conclusion}%

To resume this paper, this novel method is about clustering the membership results of a \textit{FCM} algorithm, emphasizing the intersection data. It aims to put in evidence data that are usually not in the same cluster while being very similar. The \textit{FCM} algorithm and the \textit{k-means} algorithm are used sequentially, and the final cluster consistency is evaluated with a SI metric. A filter is applied to discard the data that does not fall in any cluster intersection, allowing emphasis on intersection data. This filter uses a parameter called $\mu$, having a domain between 0 and 1. It represents the amplitude of the membership value (resulting from \textit{FCM} algorithm) the filter has to keep. The higher the value, the wider will be the size of the intersection. 
6 cases are processed to show the accuracy of this new model. This method implements several types of graphics to ensure a better explainability of the results.  

% Future work
Future works may include testing this algorithm on higher dimensionality. It would be interesting to test more systematically the scalability of this method. This method is a fruit of this fundamental research, and it would be interesting to apply it to a concrete domain in the future. Some domain ideas are mentioned in sec. \ref{sec:Discussions}.

\section{Acknowledment}%
\label{sec:Acknowledment}%
This work has been supported by the "Cellule d’expertise en robotique et intelligence artificielle" of the Cégep de Trois{-}Rivières. 
%
%\section{References}%
%\label{sec:References}%
\bibliographystyle{plain}%
\bibliography{paper}

\end{multicols}%
\end{document}